\documentclass[conference]{IEEEtran}
\IEEEoverridecommandlockouts
\usepackage{cite}
\usepackage{amsmath,amssymb,amsfonts}
\usepackage{algorithmic}
\usepackage{graphicx}
\usepackage{todonotes}
\usepackage{textcomp}
\usepackage{xcolor}
\usepackage{subcaption}
\usepackage{url}
\usepackage{booktabs}
\usepackage{hyperref}
\usepackage{svg}
\usepackage{microtype}
\usepackage{multirow}
\usepackage{amsthm}
\usepackage{mathtools}
\usepackage[ruled,vlined,linesnumbered]{algorithm2e}

\usepackage{pdflscape}
\usepackage{array}
\usepackage{longtable}

\theoremstyle{definition}
\newtheorem{definition}{Definition}[section]

\def\BibTeX{{\rm B\kern-.05em{\sc i\kern-.025em b}\kern-.08em
    T\kern-.1667em\lower.7ex\hbox{E}\kern-.125emX}}
    
\begin{document}

\title{Time-critical and confidence-based abstraction dropping methods\\
}


\author{
\IEEEauthorblockN{Robin Schmöcker}
\IEEEauthorblockA{\textit{Faculty of EECS} \\
\textit{Leibniz University Hannover}\\
Hannover, Germany \\
schmoecker@tnt.uni-hannover.de}
\and
\IEEEauthorblockN{Lennart Kampmann}
\IEEEauthorblockA{\textit{Faculty of EECS} \\
\textit{Leibniz University Hannover}\\
Hannover, Germany}

\and 
\IEEEauthorblockN{Alexander Dockhorn}
\IEEEauthorblockA{\textit{Faculty of EECS} \\
\textit{Leibniz University Hannover}\\
Hannover, Germany \\
dockhorn@tnt.uni-hannover.de}
}

\maketitle

\begin{abstract}
 One paradigm of Monte Carlo Tree Search (MCTS) improvements is to build and use state and/or action abstractions during the tree search. Non-exact abstractions, however, introduce an approximation error making convergence to the optimal action in the abstract space impossible. Hence, as proposed as a component of Elastic Monte Carlo Tree Search by Xu et al., abstraction algorithms should eventually drop the abstraction. In this paper, we propose two novel abstraction dropping schemes, namely OGA-IAAD and OGA-CAD which can yield clear performance improvements whilst being safe in the sense that the dropping never causes any notable performance degradations contrary to Xu's dropping method. OGA-IAAD is designed for time critical settings while OGA-CAD is designed to improve the MCTS performance with the same number of iterations.
\end{abstract}


\section{Introduction}
\label{sec:intro}
Monte Carlo Tree Search (MCTS) \cite{BrownePWLCRTPSC12} is a popular on-the-fly applicable decision-making algorithm. In contrast to other tree searches like MiniMax or MaxiMax, MCTS is capable of dealing with environments with large action spaces whilst requiring only a simulator of an environment. 

Due to its popularity and effectiveness, a lot of research effort is dedicated to improving MCTS. One such research area is using abstractions that aim at reducing the MCTS search space by grouping states and actions in the current MCTS search tree to enable an intra-layer information flow \cite{uctJiang,AnandGMS15,OGAUCT}. Non-exact abstractions, however, introduce an approximation error making convergence to the optimal action in the abstract space impossible. Recently, Xu et al. \cite{EMCTSXu} proposed a new paradigm for MCTS-based abstraction algorithms, which is to drop/abandon the abstraction mid-search to guarantee convergence to the optimal action. Another reason one might want to drop an abstraction is when it has no impact on the performance but its usage and calculation only hampers the runtime.

In this paper, we further investigate this technique and propose two novel dropping abstraction methods as well as provide a concrete theoretical setting in which dropping schemes provably yield performance increases.
The contributions of this paper can be summarized as follows:

\noindent \textbf{1)} We generalize On-The-Go abstractions in Upper Confidence Trees (OGA-UCT) \cite{OGAUCT} to non-exact abstractions which we call $(\varepsilon_a,\varepsilon_t)$-OGA. OGA-UCT is a technique that enables ASAP abstractions \cite{AnandGMS15} to be updated after every MCTS iteration.

\noindent \textbf{2)} We propose \textbf{OGA-IAAD}, an OGA-UCT extension that significantly reduces its runtime whilst keeping the same performance.

\noindent \textbf{3)} We propose \textbf{OGA-CAD}, an abstraction dropping scheme that operates on a node level basis and dynamically drops an abstraction based on the benefits it provides for estimating a node's true Q value. 

\noindent \textbf{4)} We reinvestigate a dropping scheme introduced by Xu et al. \cite{EMCTSXu} on stochastic Markov Decision Processes (MDP) and show that the improvements found for Strategy games cannot be reproduced in this setting.

The paper is structured as follows. Firstly, in \textbf{Section} \ref{sec:related_work} we give an overview of other abstraction algorithms and abstraction dropping methods. Then, in \textbf{Section} \ref{sec:foundations} we lay the theoretical groundwork for this paper, in particular, we formulate the abstraction framework on which we evaluate dropping methods. In \textbf{Section} \ref{sec:experiment_setup} we describe our experiment setup that applies to all experiments in this paper. The subsequent main body is split into two parts. In the first part in \textbf{Section} \ref{sec:dropping_for_time} we will introduce and evaluate OGA-IAAD, a dropping method for time-critical applications. In the second part, in \textbf{Section} \ref{sec:dropping_for_performance}, we consider abstraction-dropping methods aimed at increasing the final performance. In particular, we investigate Xu's dropping method on stochastic MDPs as well as propose and evaluate a new dropping method named OGA-CAD. At the end, in \textbf{Section} \ref{sec:future_work} we briefly summarise our findings and provide an outlook for future work. 

\section{Related Work}
\label{sec:related_work}
\subsection{MCTS-based abstractions}
Firstly, we will give an overview of MCTS-based abstraction algorithms as these are almost exclusively used in this paper.

Jiang et al. \cite{uctJiang} were the first to propose a technique to automatically detect state abstractions in parallel to running a tree search. In regular intervals, they pause MCTS (more concretely UCT \cite{KocsisS06}, but we will use UCT and MCTS synonymously) and group states within a layer when each action is pairwise approximately equivalent in the sense that their immediate reward differences as well as their transition probability differences to the node groups of the subsequent layer are below a threshold. To be able to detect any abstractions at all, they optimistically group all partially explored nodes within a layer and use a directed acyclic graph (DAG), allowing different state-action pairs to have the same successors, which is the basis for the abstraction build-up. Though the authors did not name this technique themselves, others refer to it as Abstraction of States UCT (AS-UCT) \cite{AnandGMS15}.
The computed abstraction is only used in the tree policy by improving the UCB value where instead of an action's true visits and values, one instead inserts the sum of visits and values of all corresponding actions of nodes in the same abstract node.

AS-UCT can be improved by using different grouping conditions that allow for the detection of more symmetries, for example, 
by grouping states if, for each action in a node, there is at least one equivalent action in the other node and vice versa, a condition introduced by Ravindran et al. \cite{ravindran2004approximate}. Furthermore, one can abstract nodes and actions independently. Though they are primarily state abstractions, one can also implicitly view AS-UCT as an action abstractions; however, two actions can only ever be abstracted if their parents are in the same abstract node.
These ideas are combined in Abstraction of State-Action Pairs in UCT (ASAP-UCT) \cite{AnandGMS15}, which was also proposed by Anand et al. 

The successor of ASAP-UCT is called On the Go Abstractions in UCT (OGA-UCT) \cite{OGAUCT} which improves the runtime and accuracy of ASAP-UCT by recomputing the abstraction only for frequently visited nodes thus ensuring the information contained in the abstraction does not lag behind the current search tree.

The ASAP abstraction framework is exact (i.e., the optimal policy is invariant under ASAP abstractions) 
and when approximated in MCTS, it introduces almost no abstraction errors (the only source for errors is missing successors of state-action-pairs). This makes it unsuitable for abstraction dropping if one wants to only enhance the performance and not reduce the runtime. This exactness, however, comes with the downside that many correct abstractions are not detected.

\subsection{Abstraction dropping}
Elastic Monte Carlo Tree Search (EMCTS) was proposed by Xu et al. \cite{EMCTSXu}, which was the first algorithm to explicitly drop/abandon the abstraction during MCTS. EMCTS first finds and utilises state abstractions in parallel to MCTS using the approximate-homomorphism framework exactly as done by Jiang et al. \cite{uctJiang} with their AS-UCT technique. At some fixed point, however, EMCTS drops the abstraction to continue the search in the original tree. 

Hostetler et al. \cite{HostetlerFD15} introduced PARSS, another state abstraction algorithm, which uses Forward Search Sparse Sampling (FSSS) \cite{WalshGL10}, a Sparse Sampling based tree search. PARSS constructs its abstraction by initially grouping all successors of each state-action pair. As the search progresses this coarse abstraction is refined by repeatedly splitting abstract nodes in half thus guaranteeing convergence to the original states in the limit. This splitting process can also be viewed as abstraction dropping. The abstraction dropping in PARSS is a rather steady dropping process than an instantaneous one like the one that EMCTS performs.

Though not fully domain-independent, but still widely applicable, Sokota et al. \cite{SokotaHAK21} proposed a method which we refer to as WIRSA (MCTS with iteratively refining state abstractions). Their idea is to replace sampled state-action pair successors during the tree policy with an already sampled successor if their distance (given some state-distance function) is less than some threshold $\varepsilon$. This threshold $\varepsilon$ is dependent on the iteration number and converges to zero, hence WIRSA can also be viewed, like PARSS, as performing a steady dropping process.

\section{Foundations}
\label{sec:foundations}
We use finite MDPs \cite{sutton2018reinforcement} as the model for sequential, perfect-information decision-making tasks. We use $\Delta(X)$ to denote the probability simplex of a finite, non-empty set $X$.
\begin{definition}
    An \textit{MDP} is a 6-tuple $(S,\mu_0,A_f,\mathbb{P}, R, T)$ where the components are as follows:
    \begin{itemize}
        \item $S \neq \emptyset$ is the finite set of states.
        \item $\mu_0 \in \Delta(S)$ is the probability distribution for the initial state.
        \item $A_f\colon S \mapsto A$ maps each state $s$ to the available actions $\emptyset \neq A_f(s) \subseteq A$ at state $s$ where $|A| < \infty$.
        \item $\mathbb{P}\colon S \times A \mapsto \Delta(S )$ is the stochastic transition function where we use $\mathbb{P}(s^{\prime} |\: s,a)$ to denote the probability of transitioning from $s \in S$ to $s^{\prime} \in S$ after taking action $a \in A_f(s)$ in $s$.
        \item $R \colon S \times A \mapsto \mathbb{R}$ is the reward function.
        \item $T \subseteq S$ is the (possibly empty) set of terminal states.
    \end{itemize}
We denote the set of state-action pairs by \mbox{$P \coloneqq \{(s,a)\: | \: s \in S, a \in A_f(s)\}$}.
\end{definition}

Next, we will define a parametrized abstraction framework called $\boldsymbol{(\varepsilon_a,\varepsilon_t)}$\textbf{-ASAP} that extends the ASAP framework by Anand et al. \cite{AnandGMS15} by allowing approximation errors. In general, $(\varepsilon_a,\varepsilon_t)$-ASAP will detect more correct abstractions than ASAP at the cost of an increase in faulty abstractions. This can be beneficial to the performance as we will later see and the existence of erroneous abstractions is a necessary condition for abstraction to have any benefit in the first place. Anand et al. \cite{OGAUCT} themselves already introduced a method to increase the abstraction rate in what they called \textit{pruned OGA-UCT} which simply ignores all state-action pair successors during the abstraction build up whose transition probability is below a threshold. For this paper, we chose to introduce and use $(\varepsilon_a,\varepsilon_t)$-ASAP instead as it also allows errors in the immediate reward and is more flexible as no pruned OGA threshold can guarantee that all states and state-action pair are grouped.

The $(\varepsilon_a,\varepsilon_t)$-ASAP definition is recursive, hence, we will first show how one can obtain a state equivalence relation from a state-action pair equivalence relation and vice versa, a state-action pair equivalence relation from a state equivalence relation.
Let $M = (S,\mu_0,A_f,\mathbb{P}, R, T)$ be an MDP, $0 \leq \varepsilon_a$, \mbox{$0 \leq \varepsilon_t \leq 2$}, and $\mathcal{E} \subseteq S \times S$ be an equivalence relation. We can define a reflexive and symmetric (but not necessarily transitive) relation $\approx_{\mathcal{H}} \subseteq P \times P$ for two state-action pairs $(s_1,a_1),(s_2,a_2) \in P \times P$ as
\begin{equation}
    \begin{aligned}
        (s_1,a_1) \approx_{\mathcal{H}} (s_2,a_2) \iff \quad | R(s_1,a_1) - R(s_2,a_2) | 
        & \leq \varepsilon_a \\
        \quad \textrm{and } F \coloneqq \sum \limits_{x \in \mathcal{X}} \bigg| \sum \limits_{s^{\prime} \in x} 
        \mathbb{P}(s^{\prime}|\: s_1,a_1) - \mathbb{P}(s^{\prime}|\: s_2,a_2) \bigg| 
       &  \leq \varepsilon_t.
    \end{aligned}
\end{equation}
where $\mathcal{X}$ is the set of equivalence classes of $\mathcal{E}$.
Now, we can pick some equivalence relation $\mathcal{H}^{\prime} \subseteq \approx_{\mathcal{H}} \subseteq  P \times P$ by removing elements from $\approx_{\mathcal{H}}$. Note that if $\varepsilon_a = \varepsilon_t = 0$, then $\mathcal{H}^{\prime} = \approx_{\mathcal{H}}$ is already an equivalence relation, as is the case for ASAP-UCT and OGA-UCT. In the supplementary materials Section \ref{sec:non_exact_oga} (the supplementary materials are provided in the GitHub repository accompanying this project which is available at: \url{https://github.com/codebro634/AbsDropping.git}), we describe how our OGA-UCT extension to the $(\varepsilon_a,\varepsilon_t)$-ASAP framework called $(\varepsilon_a,\varepsilon_t)$-OGA chooses $\mathcal{H}^{\prime}$ for the case that $\varepsilon_a > 0$ or $\varepsilon_t > 0$. For $\varepsilon_a=\varepsilon_t=0$, $(\varepsilon_a,\varepsilon_t)$-OGA is equivalent to standard OGA-UCT.

Next, we can define an equivalence relation $\mathcal{E}^{\prime} \subseteq S \times S$ based on $\mathcal{H}^{\prime}$ for two $s_1,s_2 \in S$ as 
\begin{equation}
    \begin{aligned}
   & (s_1,s_2) \in \mathcal{E}^{\prime}  \iff \\
        &\forall a_1 \in A_f(s_1) \, \exists a_2 \in A_f(s_2):  
        ((s_1,a_1),(s_2,a_2)) \in \mathcal{H}^{\prime} \\
        &\forall a_2 \in A_f(s_2) \, \exists a_1 \in A_f(s_1):  
        ((s_1,a_1),(s_2,a_2)) \in \mathcal{H}^{\prime}.
    \end{aligned}
\end{equation}

\begin{definition}
    We call $(\mathcal{H},\mathcal{E})$ an $(\varepsilon_a,\varepsilon_t)$\textit{-ASAP} abstraction if they are obtained from $M$ by repeatedly applying the above-mentioned constructions (i.e. state abstraction from a state-action pair abstraction and vice versa) in order, starting from some state equivalence relation. One choice for this initial relation is to put all terminal states in the same equivalence class, and all other states are their own equivalence class of size 1. \textit{(0,0)-ASAP is equivalent to the ASAP framework introduced by Anand et al. \cite{AnandGMS15}.}

    We call such an abstraction \textit{converged} if it is invariant under the above-mentioned construction steps.
\end{definition}

Repeatedly constructing $(\varepsilon_a,\varepsilon_t)$-ASAP abstractions until convergence is infeasible and such a computation would significantly hamper the runtime. Hence both ASAP-UCT \cite{AnandGMS15} and OGA-UCT \cite{OGAUCT} which were introduced in the related work section, build an $(0,0)$-ASAP like abstraction on the \textbf{local-layered MDP} (or finite horizon MDP) rooted at the state $s_d$ where the decision has to be made that differs only to an $(0,0)$-ASAP abstraction in that non-fully-expanded nodes are never grouped with expanded ones. The state space of a layered MDP is $S \times \{0,\dots,h\}$ where $h$ is the horizon and if $(s,n)$ is a successor state of $(s^{\prime},n^{\prime})$, then $n = n^{\prime} +1$ and any initial state has $n=0$. $n$ can be understood as the number of steps taken. A local-layered MDP rooted at $s_d$ is a layered MDP but with its states, actions, and possible state-action-pair-successors restricted to those present in the current search tree.

In local-layered MDPs, a converged $(\varepsilon_a,\varepsilon_t)$-ASAP abstraction can be efficiently computed with dynamic programming, where one requires only the abstraction of the previous layer to compute the abstractions for the next. The initial state equivalence relation may group all terminal states of the same layer and put all non-fully-expanded nodes of the same layer in their own equivalence class (we refer to this as \textit{non\_fully\_expanded\_abs} $ = $ \textit{single}) or aggressively clump them together into one abstract node (\textit{non\_fully\_expanded\_abs} $ = $ \textit{group}). The remaining nodes are put into their own equivalence class.   

Like AS-UCT, both OGA-UCT (and our $(\varepsilon_a,\varepsilon_t)$-OGA) and ASAP-UCT utilise the abstractions only to enhance the UCB formula during the tree policy, where they use the aggregate visits and values of all original nodes inside the abstract node of the current node.

\section{Experiment setup}
\label{sec:experiment_setup}
In this section, we describe the general experiment setup that applies to both of the subsequent sections.

\textbf{Algorithms:}
We will be using $(\varepsilon_a,\varepsilon_t)$-OGA in the next sections and since $(\varepsilon_a,\varepsilon_t)$-OGA performs tree search on a DAG, we will also perform standard MCTS on a DAG (i.e. allow two state-action pairs to share successor state nodes), whenever we compare it to an OGA variant. Furthermore, since the environments we will later run MCTS on have vastly different reward scales, we will use a dynamic, scale-independent exploration factor which 
has the form $\lambda \cdot \sigma$ where $\sigma$ is the standard deviation of the Q values of all nodes in the search tree and $\lambda \in \mathbb{R}^+$. We use this for both MCTS and $(\varepsilon_a,\varepsilon_t)$-OGA. Whenever we refer to standard MCTS, we implicitly mean MCTS using these two techniques (dynamic exploration factor + DAG).

\textbf{Problem models:}
For this paper, we ran our experiments on a variety of MDPs, most of which are from the International Probabilistic Planning Conference \cite{grzes2014ippc} and are widely used in the abstraction algorithm literature.
We ran all of our experiments on the finite horizon versions of the considered MDPs with a horizon length of 50 and a discount factor $\gamma=1$. If the reader is not familiar with any of the domains we used for the experiments, we provide a brief description for each MDP in the supplementary materials in Section~\ref{sec:problem_descriptions}. For an exact description of the problems, we refer to our GitHub repository available at \url{https://github.com/codebro634/AbsDropping.git}.

\textbf{Evaluation:}
Each data point that we denote in the remaining sections of this paper (e.g. agent returns) is the average of at least 1000 runs. Whenever we denote a confidence interval for a data point then this is always a bootstrapped confidence interval with a confidence level of 99\%. If a data point appears in any figure its corresponding error bars denote this 99\% confidence interval.

\textbf{Reproducibility:}
For the time-critical experiments, we measured the runtime on a single core of an Intel(R) Core(TM) i5-9600K CPU @ 3.70GHz. Our code was compiled with g++ version 13.1.0 using the -O3 flag (i.e. aggressive optimization). For reproducibility purposes, we seeded all components of our program that require RNG. For each data point that was acquired, we used the seed 42 for the RNG engine.

\section{Abstraction dropping for computational efficiency}
\label{sec:dropping_for_time}
\subsection{Method}
Let us assume OGA-UCT has to be run under time-critical circumstances which is the main motivation behind the development of OGA-UCT in the first place, as it only speeds up the computation of ASAP abstractions.

Abstraction computation in OGA-UCT can cause a significant runtime overhead. Even if OGA-UCT hasn't found any abstractions, it continues to search for them, which slows down the computation.
An example where OGA-UCT finds next to no abstractions is the Game of Life IPPC problem which features a high stochastic branching factor. In fact, for each action, there are $2^n$ possible successor states if $n$ is the width of the map. Consequently, once two actions have accumulated only a few visits, it is practically impossible for both to have sampled the same set of outcomes, a necessary condition for OGA-UCT to abstract these.

This overhead can be mitigated if one stops the abstraction computation when one is certain that the abstractions will not have a significant impact on the performance which is implied when few or no abstractions are found. We use a proxy quantity to quantify the abstraction impact. The proxy we use is the \textbf{compression rate} $C \in [1,\infty)$ which is defined as 
        \mbox{$C \coloneqq \max (C_s,C_a)$}
where $C_s$ is ratio of total state nodes divided by the total number of abstract state nodes and $C_s$ is the ratio of total Q state nodes divided by the number of abstract Q state nodes (a Q state node in an MCTS tree represents a state-action pair). Note that Xu et al. \cite{EMCTSXu} also defined a compression rate that is equal to $C_s$ as they considered a state-only abstraction.

We propose an OGA-UCT plugin to reduce its runtime while keeping the performance which operates as follows. Once an iteration threshold $\tau \in [0,1]$ (denoted as a ratio of the total number of iterations) is reached, one checks every $n_{check}$ iterations whether $C$ is below some threshold $\hat{C} \in [1,\infty)$. If so, then the abstraction computation is stopped until the end of the tree search and one regresses to using the original visits and values for the action selection in the tree policy. For abbreviation purposes, we call this OGA-UCT modification, OGA-\textbf{I}mpact-\textbf{A}ware-\textbf{A}bstraction-\textbf{D}ropping (OGA-IAAD). Note that OGA-IAAD and OGA-UCT are equivalent if either $\tau = 1$ or $\hat{C} =1$.
\subsection{Experiments}
\label{sec:droppin_for_time_exp}
\begin{figure*} 
    \centering

    \begin{subfigure}[b]{0.24\textwidth}
        \centering
        \includegraphics[width=\textwidth]{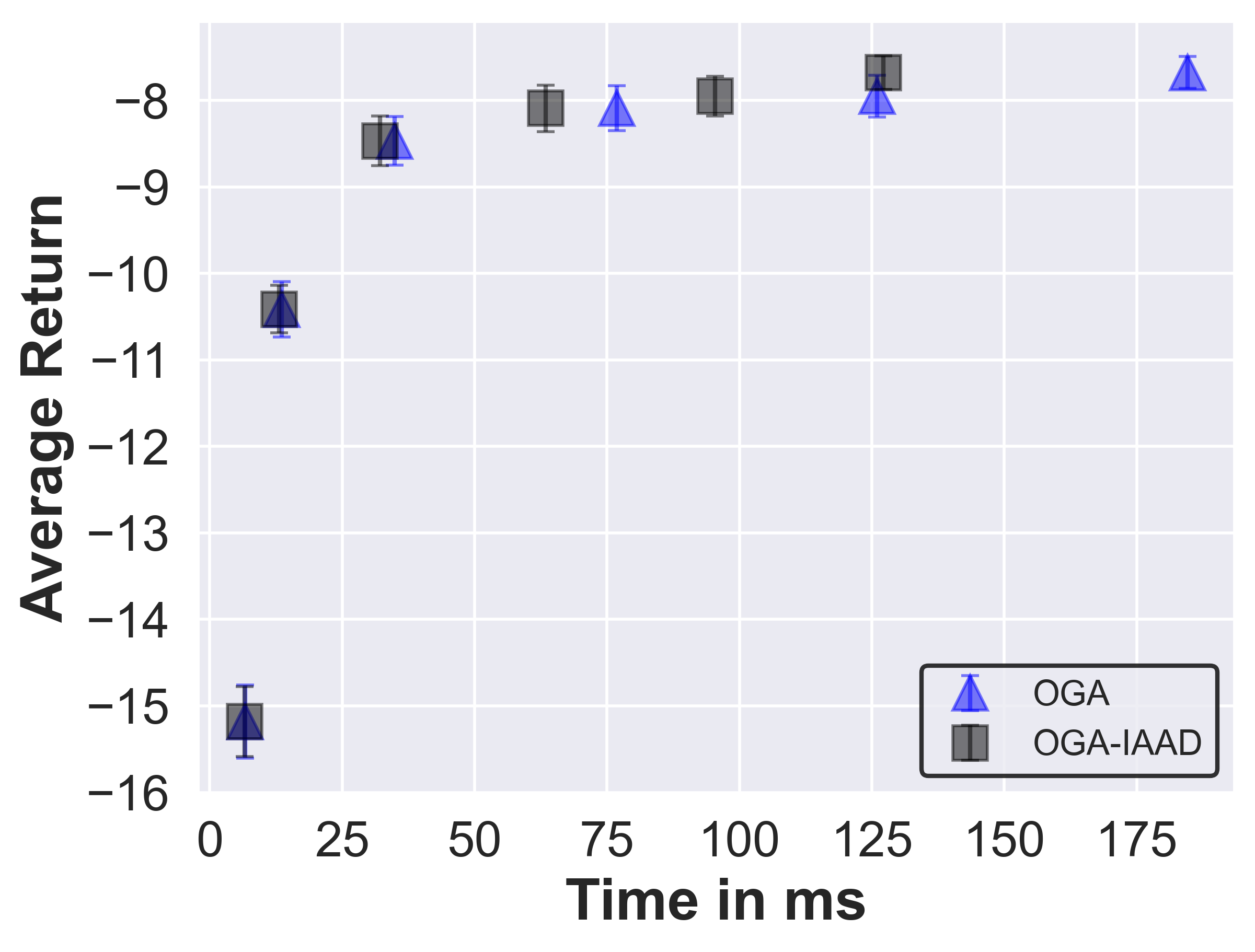}
        \caption{Earth Observation}
    \end{subfigure}
    \hfill
    \begin{subfigure}[b]{0.24\textwidth}
        \centering
        \includegraphics[width=\textwidth]{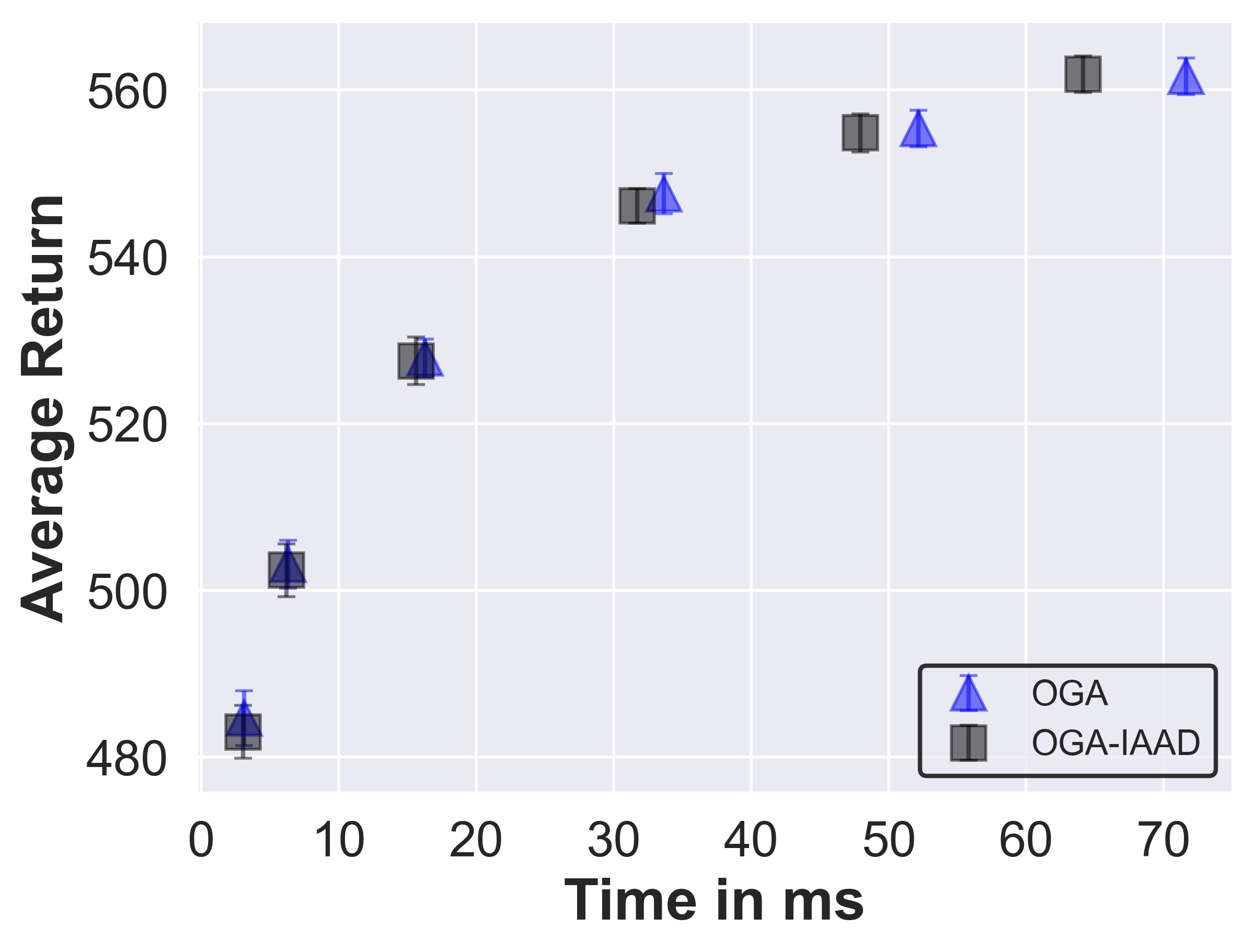}
        \caption{Game of Life}
        
    \end{subfigure}
    \hfill
    \begin{subfigure}[b]{0.24\textwidth}
        \centering
        \includegraphics[width=\textwidth]{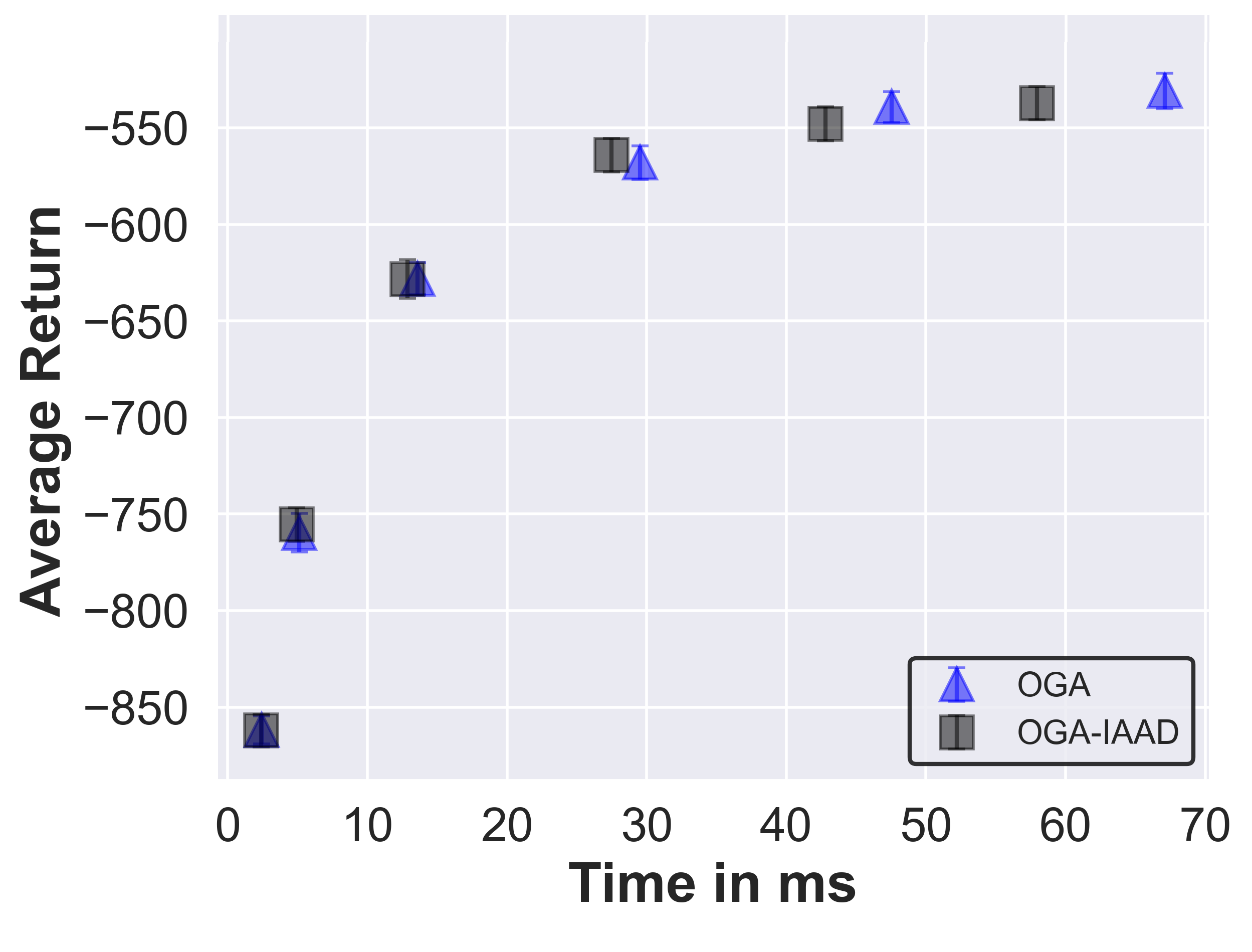}
        \caption{Tamarisk}
         
    \end{subfigure}
   \hfill
    \begin{subfigure}[b]{0.24\textwidth}
    \centering
    \includegraphics[width=\textwidth]{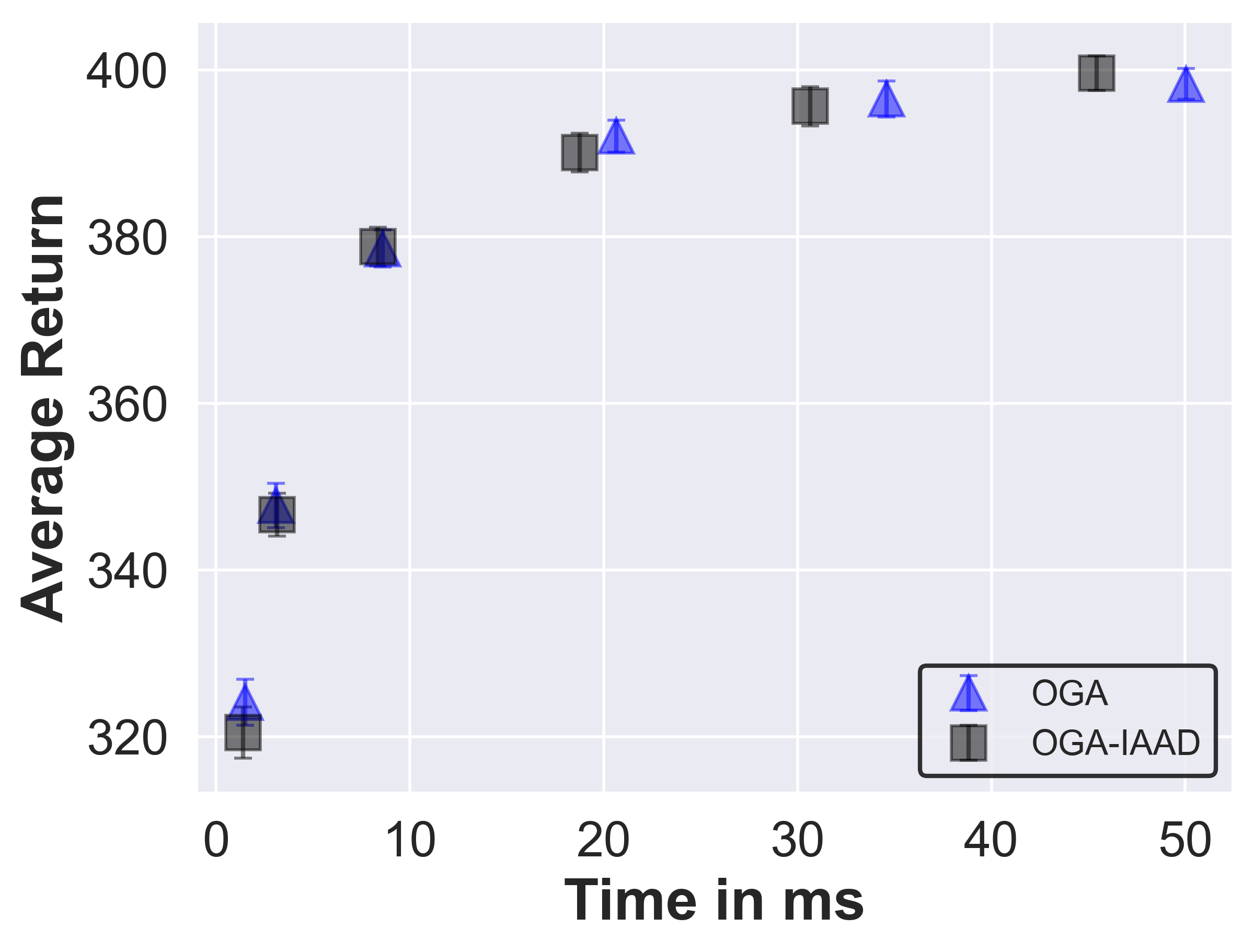}
    \caption{SysAdmin}
        
    \end{subfigure}

    \caption{Performance comparison of OGA-IAAD with OGA. The x-axis is the measured runtime for each of the above-mentioned iteration counts. Hence, two neighboring data points on the x-axis correspond to two neighboring iteration counts $n$. E.g. the leftmost data points correspond to $n=100$ and the rightmost one to $n=2000$.}
    \label{fig:oga_sd}
\end{figure*}

Next, we will experimentally investigate OGA-IAAD. To accomplish this, we ran both OGA and OGA-IAAD
for $n \in \{100,200,500,1000,1500,2000\}$ iterations with an exploration factor $\lambda=2$, a drop-check threshold $\tau = 0.25$, a compression rate threshold $\hat{C} \in \{1.01\}$, and a recency count $K = 3 $ (the frequency of node visits until the abstraction is recomputed), the value that Anand et al. \cite{AnandGMS15} determined to have the best runtime-performance tradeoff, and the check frequency $n_{check} = 10$.
Besides measuring the agents' final performance, we also measured their average runtime. Figure~\ref{fig:oga_sd} visualizes the time-dependent performance graphs for a varying iteration count on a subset of problem models with a high stochastic branching factor. Clearly, OGA-IAAD yields a clear runtime improvement whilst not losing performance. Furthermore, Table \ref{tab:oga_sd} demonstrates that OGA-IAAD does not cause performance degradation in environments where OGA does detect abstractions. The only clear performance degradations due to this dropping scheme were sometimes present in the 100-iteration setting for environments with at least moderately sized action spaces such as Academic Advising. This is because the check for abstraction dropping occurs at a time point where there weren't even enough iterations to trigger any action's abstraction update, hence the abstraction that would be in a future update is not present at the abstraction drop check point.

\begin{table}[h]\centering

\caption{The performances of both OGA and OGA-IAAD for 2000 iterations on several problem domains. The time columns denote the average decision time per action in milliseconds.}
\scalebox{0.8}{
\setlength{\tabcolsep}{1mm}\begin{tabular}{c| c c|c c}
\toprule
\multirow{2}{*}{Domain} & \multicolumn{2}{c|}{Ret} & \multicolumn{2}{c}{Time} \\
 &OGA-IAAD & OGA-UCT & OGA-IAAD & OGA-UCT \\
\midrule
Academic Advising & $\boldsymbol{-62.6 \pm 0.9}$ & $-63.0 \pm 0.9$ & $39.8 \pm 1.1$ & $\boldsymbol{39.4 \pm 1.1}$\\
Earth Observation & $\boldsymbol{-7.68 \pm 0.20}$ & $\boldsymbol{-7.68 \pm 0.21}$ & $\boldsymbol{127.2 \pm 3.3}$ & $184.6 \pm 3.2$\\
Game of Life & $\boldsymbol{561.9 \pm 2.3}$ & $561.7 \pm 2.3$ & $\boldsymbol{64.1 \pm 1.6}$ & $71.6 \pm 1.7$\\
Manufacturer & $\boldsymbol{-1146.3 \pm 9.6}$ & $-1152.2 \pm 9.8$ & $\boldsymbol{205.0 \pm 5.0}$ & $218.1 \pm 5.1$\\
Navigation & $-16.3 \pm 0.5$ & $\boldsymbol{-16.1 \pm 0.5}$ & $42.9 \pm 1.7$ & $\boldsymbol{42.3 \pm 2.0}$\\
Cooperative Recon & $\boldsymbol{14.6 \pm 0.2}$ & $\boldsymbol{14.6 \pm 0.2}$ & $166.2 \pm 4.4$ & $\boldsymbol{165.9 \pm 4.4}$\\
Racetrack & $\boldsymbol{-8.49 \pm 0.04}$ & $\boldsymbol{-8.49 \pm 0.04}$ & $\boldsymbol{20.5 \pm 1.5}$ & $\boldsymbol{20.5 \pm 1.5}$\\
SysAdmin & $\boldsymbol{399.6 \pm 1.9}$ & $398.3 \pm 2.1$ & $\boldsymbol{45.4 \pm 0.8}$ & $50.0 \pm 0.8$\\
Skills Teaching & $34.8 \pm 8.7$ & $\boldsymbol{35.2 \pm 9.1}$ & $111.2 \pm 2.1$ & $\boldsymbol{111.1 \pm 1.9}$\\
Sailing Wind & $-61.4 \pm 1.4$ & $\boldsymbol{-60.5 \pm 1.4}$ & $\boldsymbol{53.7 \pm 1.2}$ & $56.9 \pm 1.3$\\
Tamarisk & $-537.2 \pm 9.2$ & $\boldsymbol{-530.9 \pm 8.7}$ & $\boldsymbol{57.9 \pm 1.1}$ & $67.1 \pm 1.2$\\
Traffic & $\boldsymbol{-13.2 \pm 0.3}$ & $-13.3 \pm 0.3$ & $\boldsymbol{71.0 \pm 1.5}$ & $73.0 \pm 1.5$\\
Triangle Tireworld & $\boldsymbol{81.5 \pm 1.1}$ & $\boldsymbol{81.5 \pm 1.1}$ & $\boldsymbol{51.4 \pm 2.6}$ & $\boldsymbol{51.4 \pm 2.6}$\\
\bottomrule
\end{tabular}}

\label{tab:oga_sd}
\end{table}

\section{Abstraction dropping for performance gain}
\label{sec:dropping_for_performance}

\subsection{Eliminating confounders}
\label{sec:confounders}

This section has two aims. The first one is to propose and evaluate an improvement to the dropping scheme introduced by Xu et al. \cite{EMCTSXu}, which is to drop the entire abstraction once an iteration number threshold has been reached. From hereon, we refer to Xu's dropping technique as \textbf{I}teration-based-\textbf{S}imultaneous-\textbf{D}ropping (ISD).
Secondly, we 
reanalyze the abstraction-dropping method by Xu et al. on stochastic MDPs as Xu et al. originally investigated abstraction dropping only on Strategy games that are deterministic and multiplayer; hence, they are qualitatively different from stochastic MDPs, thus justifying an analysis on these domains too.
In particular, the parameter $\varepsilon_t$ is irrelevant for deterministic games, even though it is a core component of the ASAP framework.

Since the underlying abstraction algorithm will be computationally expensive, the basic assumption of this section is that we are in a setting where one wants to improve performance with the same number of MCTS iterations, because calls to the environment model are assumed to be costly. 
Since our focus is on the impact of abstraction-dropping on performance, it is important to first address potential confounding factors.

\noindent\textbf{1) Exploration factor}: The number of abstractions implicitly control exploration. For example, the trivial abstraction where all states and actions of the same depth are abstracted together is equivalent to using an exploration factor of $\infty$. Hence, we first determine the optimal exploration constant of MCTS for each environment-iterations combination. For all subsequent experiments of $(\varepsilon_a,\varepsilon_t)$-OGA using the same iteration count and environment, we use the previously determined constant to ensure performance cannot be gained by varying the exploration strategy. The data for this can be found in the supplementary materials Tab.~\ref{tab:params}.

\noindent\textbf{2) Abstraction delay}: In this work, we only want to focus on the approximation error of the abstraction framework itself and not on the error of the abstraction lagging behind the information contained in the tree search. To make investigating this feasible, we used $(\varepsilon_a,\varepsilon_t)$-OGA instead of OGA-UCT (i.e. $\varepsilon_a=\varepsilon_t=0$) i.e. non-exact abstractions, a necessary requirement for abstraction dropping to have any beneficial impact. For details of our method, we refer to the supplementary materials Section \ref{sec:non_exact_oga}.

\noindent\textbf{3) Worse than MCTS abstractions}: We also want to eliminate the possibility that the abstraction algorithm is simply worse than standard MCTS and the performance gain only comes from the fact that we abandon the worsening of MCTS. Therefore, we will also always compare performances to MCTS.

\noindent\textbf{4) Decoupling abstraction improvements from dropping improvements}: The EMCTS algorithm combines both abstractions and abstraction dropping. However, to distill the potential improvements coming from the abstraction dropping itself, a comparison with the underlying abstraction algorithm but without abstraction dropping has to be made.

\subsection{Why abstraction dropping can be beneficial}
\label{sec:improvements}
\begin{figure}
    \centering
    \includegraphics[width=0.3\textwidth]{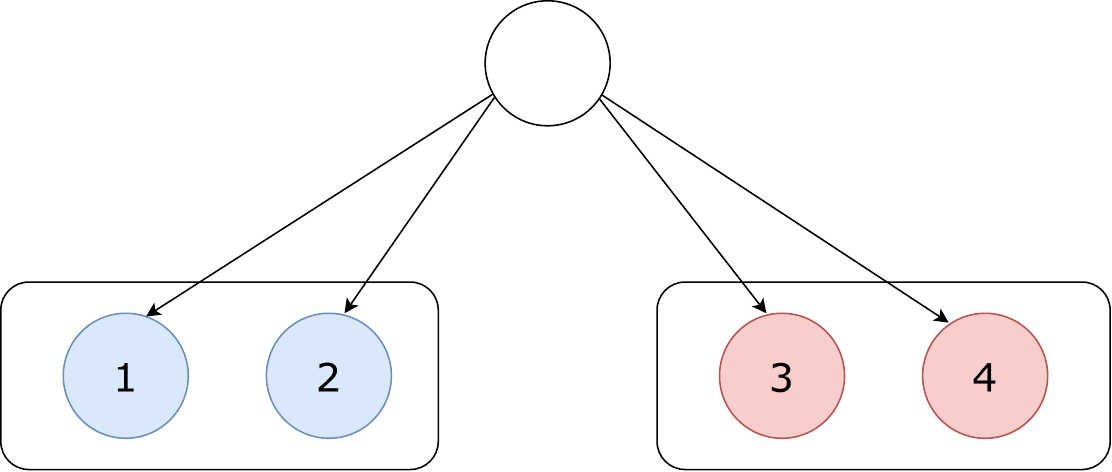}
   \caption{Assume that MCTS is run on the following depth-1 game tree and that we abstract the actions leading to nodes 1 and 2 together as well those leading to nodes 3 and 4. Furthermore, assume that all actions have a noisy immediate reward, with no two actions having the same expected payoff, action 4 having the highest expected payoff, and action 3 has the second highest expected payoff. Conducting MCTS using these abstractions will converge to choosing either action 3 or 4 more quickly than if no abstractions were used. However, with these abstractions, it is not possible to detect that action 4 is better than 3. Hence, it would be beneficial to drop the abstraction once the search has converged to 3 and 4.
   }
    \label{fig:abs_drop}
\end{figure}

Abstraction dropping might seem counterintuitive at first. If abstraction usage clearly improves performance over standard MCTS, why would we abandon the abstraction? The reason is that running MCTS on an abstract tree does not guarantee convergence to the optimal action(s) if the abstraction is not exact. Hence, dropping the abstraction at some point can guarantee optimality and can be seen as switching from a search on a higher level to a refinement process, as mentioned by Xu et al. \cite{EMCTSXu}. Figure~\ref{fig:abs_drop} visualizes a concrete game tree where abstraction dropping would outperform both search without abstractions and search using abstractions until the end.

\subsection{Opportunities for improvement in abstraction dropping}
We identified four areas of improvement in Xu et al.'s dropping scheme.

\noindent\textbf{1) Overestimation bias}: The exploration term in the UCB formula quantifies potential performance gains that could be obtained with more exploitation. If one performs MCTS on a DAG without any abstractions, this exploration term is overestimated as different actions that converge to the same state (either immediately or multiple steps downstream) do not share visits but both exploit that state. Abstractions counteract this problem, which is reinstated if one uses the original visits after the abstraction drop again. 

    \noindent\textbf{2) An additional parameter}: It is unclear when to drop the abstraction. If one drops too early, one might degrade to standard MCTS, while dropping too late may leave not enough time for refinement. Hence, one introduces yet another parameter that has to be tuned.
    
  \noindent \textbf{3) No node-level sensitivity}: Another benefit of abstractions is the variance reduction of an action's Q value. Ideally, one only wants to keep the abstraction as long as the benefits of variance reduction outweigh the potential bias the abstraction introduces. However, the point in time when this happens is different for each action. Therefore, dropping everything at once cannot be optimal.

 \noindent\textbf{4) An implicit tradeoff}: The more actions an abstract action encompasses, the more virtual visits are introduced. E.g. in an abstraction with $n$ actions with $m$ visits each, each action has an abstract visit count of $n\cdot m$. Hence, abstractions result in a higher exploitation as these additional visits artificially dampen the UCB exploration term. Considering Figure~\ref{fig:abs_drop} again, this would mean that after dropping the abstraction, we remove these virtual visits, causing MCTS to revisit actions 1 and 2 more often than if we kept the virtual visits. However, keeping the virtual visits means that the differentiation between 3 and 4 is more difficult due to a lower exploration.


\subsection{Confidence-based abstraction dropping}

Next, we propose an alternative to the ISD dropping method by Xu et al. \cite{EMCTSXu} that addresses problem 3 and alleviates problem 2.

\noindent \textbf{Decoupling}: Our idea is to decouple abstraction usage and abstraction construction. In our dropping method, we never stop building the abstraction but only modify how the abstraction affects the UCB value during the tree policy. UCB is modified by dropping the abstraction on a per state-action-pair level. We use the original visits and values for any action whose abstraction has been dropped and stick to the abstract values and visits for those that aren't. Whether a state-action pair, which we refer to as Q nodes in the following, drops its abstraction is described in the following:

\noindent \textbf{Q node abstraction dropping}: The idea is to drop the abstraction for each original Q node $q$ whenever we are certain that the Q value $Q_q$ of $q$ is closer to the value under optimal play $Q^*$ of that node than the Q value $Q_{abs,q}$ of $q$'s abstract node. More concretely, using the sum of squared returns, the visits, and the sum of returns, we build an approximate confidence interval $[Q_q-r,Q_q+r] = I_q \subseteq \mathbb{R}$ for $Q^*$. We then drop the abstraction if and only if
\begin{equation}
    \frac{r}{2} < \min \big (|Q_{abs,q} - (Q_q-r)|,|Q_{abs,q} - (Q_q+r)| \big ).
\end{equation}
To be able to revert wrong drop-decisions, $q$ may start using its abstraction again if the above inequality does no longer hold.
And though this technique introduces a new parameter $p \in [0,1]$, the confidence level for the confidence interval, performances are mostly invariant to the choice of this parameter as we shall see later. 

We call the above-described method \textbf{C}onfidence-based-\textbf{A}bstraction-\textbf{D}ropping (CAD). Note that for $p=1$ OGA-CAD is equivalent to $(\varepsilon_a,\varepsilon_t)$-OGA and for $p=0$ OGA-CAD is equivalent to standard MCTS under the assumption that there are no abstract Q nodes with size $ > 1$ that all contain the same values.

\subsection{Experiments}
\label{sec:dropping_for_improv_exp}
\noindent \textbf{The impact of ISD}: First, we will experimentally compare the ISD dropping scheme of EMCTS using $(\varepsilon_a,\varepsilon_t)$-OGA (OGA-ISD) with $(\varepsilon_a,\varepsilon_t)$-OGA and standard MCTS. The exploration constant per domain is set to the optimum for standard MCTS (see supplementary materials Tab.~\ref{tab:params}). For both $(\varepsilon_a,\varepsilon_t)$-OGA and OGA-ISD, we tested the cross product of $\textit{\textrm{non\_fully\_expanded\_abs}} \in \{group,single\}$, $\varepsilon_t \in \{0,0.4,1.0,2.0\}$ and $\varepsilon_a$ with up to three environment dependent values that include $\varepsilon_a=0$, a moderate and high value for that environment. For OGA-ISD, we tested an iteration threshold $\tau \in \{0.25,0.5,0.75\}$, i.e. the point at which we drop the abstraction. 

Table~\ref{tab:oga_naive_highvar} shows the maximal performance of all parameter combinations. Not only is OGA-ISD not able to significantly improve the performance in any environment, OGA-ISD usually suffers a performance loss compared to $(\varepsilon_a,\varepsilon_t)$-OGA. In this setting, one reaches a peak performance by using an abstraction that introduces very little error (for $(\varepsilon_a,\varepsilon_t)$-OGA the best performances were reached with $\varepsilon_t \leq 0.4$, with an in general low $\varepsilon_a$, and mostly $\textit{\textrm{non\_fully\_expanded\_abs}}=0$) and keeping that abstraction instead of starting with a coarse abstraction and dropping it later.

\begin{table*}[]\centering

\caption{Average returns and 99\% bootstrap confidence interval for MCTS, $(\varepsilon_a,\varepsilon_t)$-OGA, and OGA-ISD in the \textbf{high variance setting}. The number behind OGA-ISD indicates the drop threshold $\tau$.}

\scalebox{0.75}{
\setlength{\tabcolsep}{1mm}\begin{tabular}{c|c c c c c c }
\toprule
Domain &MCTS & $(\varepsilon_a,\varepsilon_t)$-OGA & OGA-CAD & OGA-ISD 0.25 & OGA-ISD 0.5 & OGA-ISD 0.75\\
\midrule
Academic Advising & $-70.0 \pm 0.9$ & $\boldsymbol{-65.1 \pm 0.9}$ & $-66.8 \pm 0.9$ & $-68.4 \pm 1.0$ & $-67.4 \pm 0.9$ & $-68.6 \pm 1.0$\\
Cooperative Recon & $6.71 \pm 0.43$ & $14.2 \pm 0.2$ & $\boldsymbol{14.3 \pm 0.2}$ & $7.36 \pm 0.49$ & $9.44 \pm 0.45$ & $12.0 \pm 0.4$\\
Earth Observation & $-8.01 \pm 0.25$ & $-7.98 \pm 0.24$ & $-8.02 \pm 0.24$ & $-8.02 \pm 0.24$ & $\boldsymbol{-7.93 \pm 0.24}$ & $-8.13 \pm 0.27$\\
Game of Life & $549.2 \pm 2.3$ & $\boldsymbol{569.2 \pm 2.4}$ & $566.5 \pm 2.3$ & $549.7 \pm 2.3$ & $549.1 \pm 2.3$ & $545.6 \pm 2.5$\\
Manufacturer & $-1235.7 \pm 13.4$ & $-1128.4 \pm 10.6$ & $\boldsymbol{-1127.5 \pm 10.1}$ & $-1212.4 \pm 13.7$ & $-1204.0 \pm 12.2$ & $-1181.7 \pm 11.3$\\
Navigation & $-20.9 \pm 0.8$ & $-11.2 \pm 0.2$ & $\boldsymbol{-11.1 \pm 0.2}$ & $-12.5 \pm 0.3$ & $-12.3 \pm 0.3$ & $-12.6 \pm 0.2$\\
Racetrack & $\boldsymbol{-9.01 \pm 0.07}$ & $-9.10 \pm 0.04$ & $-9.12 \pm 0.04$ & $-9.04 \pm 0.07$ & $-9.07 \pm 0.06$ & $-9.21 \pm 0.05$\\
Sailing Wind & $-63.1 \pm 1.3$ & $-61.8 \pm 1.5$ & $\boldsymbol{-61.3 \pm 1.4}$ & $-62.7 \pm 1.4$ & $-62.7 \pm 1.4$ & $-63.0 \pm 1.4$\\
Skills Teaching & $27.2 \pm 8.6$ & $65.6 \pm 8.7$ & $\boldsymbol{63.2 \pm 8.3}$ & $50.9 \pm 8.8$ & $59.5 \pm 8.7$ & $65.4 \pm 9.2$\\
SysAdmin & $390.7 \pm 2.1$ & $\boldsymbol{402.2 \pm 2.2}$ & $401.1 \pm 2.3$ & $392.6 \pm 2.0$ & $394.1 \pm 2.2$ & $388.9 \pm 2.2$\\
Tamarisk & $-563.9 \pm 9.1$ & $\boldsymbol{-546.5 \pm 9.2}$ & $-551.6 \pm 9.0$ & $-559.4 \pm 9.4$ & $-556.1 \pm 9.2$ & $-563.8 \pm 9.2$\\
Traffic &  $-13.7 \pm 0.3$ & $-13.6 \pm 0.3$ & $-13.7 \pm 0.3$ & $\boldsymbol{-13.5 \pm 0.3}$ & $\boldsymbol{-13.5 \pm 0.3}$ & $\boldsymbol{-13.5 \pm 0.3}$\\
Triangle Tireworld & $79.2 \pm 1.3$ & $81.0 \pm 1.1$ & $81.4 \pm 1.0$ & $79.9 \pm 1.2$ & $80.7 \pm 1.2$ & $\boldsymbol{81.9 \pm 0.9}$\\
\bottomrule
\end{tabular}}

\label{tab:oga_naive_highvar}
\end{table*}

\noindent \textbf{OGA-CAD versus OGA-ISD}: In this section, we will investigate our previously proposed OGA-CAD dropping technique. Since the technique only drops an abstraction when it is certain of its harmfulness, OGA-CAD reaches the same peak performances as standard $(\varepsilon_a,\varepsilon_t)$-OGA. This is because in our experiment setting, Q values are in general very noisy and since the best $(\varepsilon_a,\varepsilon_t)$-OGA performances are reached with near-lossless abstractions, OGA-CAD correctly detects that these abstractions should not be dropped, contrary to what OGA-ISD does. Tab.~\ref{tab:oga_naive_highvar} shows the peak performances of OGA-CAD if we vary $\varepsilon_a, \varepsilon_t$, and $\textit{\textrm{non\_fully\_expanded\_abs}} \in \{group,single\}$, and use a drop confidence $p=0.5$. Clearly, even with the most volatile $p$ value we tested, OGA-CAD only sporadically drops abstraction because it lacks the confidence to do so.

Next, we want to show instances in which OGA-CAD does correctly drop the abstraction for a performance gain. We test whether OGA-CAD can  correctly drop a very coarse and imprecise abstraction for performance gain as well as whether it can drop finer abstractions to potentially improve the peak performance of $(\varepsilon_a,\varepsilon_t)$-OGA, which is only attained at finer abstractions. We need to modify our experimental setting slightly to reduce noise. For the following experiment, each rollout is repeated 10 times, and the average return of these rollouts is used for backpropagation. Furthermore, we reduce the rollout length to a problem-specific value (see supplementary materials Tab.~\ref{tab:params}). First, we test whether OGA-CAD can correctly and decisively drop very coarse abstractions. Tab.~\ref{tab:oga_cad_lowvar_coarse} compares the performances of $(\varepsilon_a,\varepsilon_t)$-OGA and OGA-CAD (with $p \in \{0.5,0.75,0.9\}$) and $\varepsilon_t = 2.0$, \textit{non\_fully\_expanded\_abs} $=$ \textit{group}, and $\varepsilon_a$ is set to the highest value we tested for the specific environment. Clearly, even with the highest confidence level, OGA-CAD does decisively drop the abstraction to gain a performance improvement.

\begin{table*}[]\centering

\caption{Average return and 99\% bootstrap confidence intervals for $(\varepsilon_a,\varepsilon_t)$-OGA and OGA-CAD with varying confidence levels in the \textbf{low variance setting} using coarse abstractions.}

\scalebox{0.75}{
\setlength{\tabcolsep}{1mm}\begin{tabular}{c|c c c c }
\toprule
Domain &$(\varepsilon_a,\varepsilon_t)$-OGA & OGA-CAD 0.5 & OGA-CAD 0.75 & OGA-CAD 0.9\\
\midrule
Academic Advising & $-255.8 \pm 4.3$ & $-112.9 \pm 1.8$ & $\boldsymbol{-111.6 \pm 2.1}$ & $-136.0 \pm 2.4$\\
Cooperative Recon & $-1.00 \pm 0.05$ & $2.59 \pm 0.34$ & $5.12 \pm 0.45$ & $\boldsymbol{6.20 \pm 0.46}$\\
Earth Observation & $-75.7 \pm 0.8$ & $-29.0 \pm 0.7$ & $-25.4 \pm 0.6$ & $\boldsymbol{-24.1 \pm 0.5}$\\
Game of Life & $402.6 \pm 4.2$ & $538.4 \pm 2.4$ & $\boldsymbol{543.2 \pm 2.6}$ & $521.5 \pm 2.7$\\
Manufacturer & $-4248.6 \pm 78.6$ & $\boldsymbol{-1246.7 \pm 10.5}$ & $-1364.2 \pm 11.3$ & $-1450.7 \pm 15.1$\\
Navigation & $-47.6 \pm 0.5$ & $\boldsymbol{-22.6 \pm 0.8}$ & $-28.0 \pm 0.9$ & $-31.6 \pm 1.0$\\
Racetrack & $-49.0 \pm 0.3$ & $\boldsymbol{-14.6 \pm 0.4}$ & $-16.8 \pm 0.5$ & $-21.3 \pm 0.7$\\
Sailing Wind & $-117.8 \pm 0.5$ & $\boldsymbol{-90.2 \pm 1.2}$ & $-96.1 \pm 1.0$ & $-100.9 \pm 0.9$\\
Skills Teaching & $-547.6 \pm 4.1$ & $\boldsymbol{-137.6 \pm 9.2}$ & $-188.7 \pm 9.7$ & $-232.9 \pm 10.1$\\
SysAdmin & $230.4 \pm 2.7$ & $\boldsymbol{336.9 \pm 2.4}$ & $311.9 \pm 2.5$ & $298.3 \pm 2.4$\\
Tamarisk & $-1118.6 \pm 5.5$ & $\boldsymbol{-723.0 \pm 11.7}$ & $-780.8 \pm 11.9$ & $-811.0 \pm 12.2$\\
Traffic & $-79.2 \pm 1.2$ & $\boldsymbol{-14.0 \pm 0.3}$ & $-15.5 \pm 0.4$ & $-17.5 \pm 0.4$\\
Triangle Tireworld & $-17.1 \pm 3.2$ & $\boldsymbol{58.6 \pm 3.0}$ & $56.4 \pm 3.1$ & $30.6 \pm 3.6$\\
\bottomrule
\end{tabular}}

\label{tab:oga_cad_lowvar_coarse}
\end{table*}

Next, we want to test if OGA-CAD can improve the peak performance of $(\varepsilon_a,\varepsilon_t)$-OGA.
Tab.~\ref{tab:oga_cad_lowvar} compares the maximal performances of MCTS, $(\varepsilon_a,\varepsilon_t)$-OGA, OGA-ISD, and OGA-CAD (with $p \in \{0.5,0.75,0.9\}$) across all parameter choices.
What we find is that OGA-CAD is mostly insensitive to the concrete choice of $p$ and can yield significant performance improvements in Game of Life, SysAdmin, and Skills Teaching. In the games where the peak performance is not improved, the performances of $p=0.75$ and $p=0.9$ are tightly within the error bounds.

On the contrary, OGA-ISD is volatile in the sense that there are two environments, namely Racetrack and Game of Life, where OGA-ISD yields a minor performance improvement and one environment, namely Manufacturer, where it significantly increases the peak performance, but at the same time, OGA-ISD
is highly sensitive to $\tau$, in fact, in around half of the tested environments, at least one value of $\tau$ significantly degrades the performance, which is problematic as there is not a clear value range as with OGA-CAD in which $\tau$ should be chosen, since in the extreme cases $\tau$ should be near $1$ for errorless abstraction but near $0$ for highly erroneous abstractions.

\begin{table*}[]\centering

\caption{Average return and 99\% bootstrap confidence intervals for MCTS, $(\varepsilon_a,\varepsilon_t)$-OGA, OGA-ISD, and OGA-CAD in the \textbf{low variance setting}. The number behind OGA-ISD indicates the drop threshold $\tau$, and the number behind OGA-CAD indicates the confidence threshold $p$.}

\scalebox{0.75}{
\setlength{\tabcolsep}{1mm}\begin{tabular}{c|c c c c c c c c}
\toprule
Domain &MCTS & $(\varepsilon_a,\varepsilon_t)$-OGA & OGA-CAD 0.5 & OGA-CAD 0.75 & OGA-CAD 0.9 & OGA-ISD 0.25 & OGA-ISD 0.5 & OGA-ISD 0.75\\
\midrule
Academic Advising & $-66.6 \pm 0.5$ & $\boldsymbol{-66.0 \pm 0.8}$ & $-68.3 \pm 0.9$ & $-66.6 \pm 0.9$ & $-66.4 \pm 1.0$ & $-66.2 \pm 0.9$ & $-66.6 \pm 0.9$ & $-65.4 \pm 0.9$\\
Cooperative Recon & $10.4 \pm 0.4$ & $\boldsymbol{15.8 \pm 0.3}$ & $\boldsymbol{15.8 \pm 0.3}$ & $\boldsymbol{15.8 \pm 0.3}$ & $\boldsymbol{15.8 \pm 0.2}$ & $12.7 \pm 0.4$ & $15.0 \pm 0.3$ & $15.6 \pm 0.3$\\
Earth Observation & $-7.72 \pm 0.23$ & $\boldsymbol{-7.55 \pm 0.20}$ & $-7.64 \pm 0.23$ & $-7.62 \pm 0.19$ & $-7.59 \pm 0.20$ & $-7.63 \pm 0.19$ & $-7.66 \pm 0.20$ & $-7.63 \pm 0.22$\\
Game of Life & $565.4 \pm 1.5$ & $566.8 \pm 2.1$ & $568.2 \pm 2.4$ & $\boldsymbol{570.5 \pm 2.4}$ & $566.5 \pm 2.2$ & $567.1 \pm 2.3$ & $570.1 \pm 2.3$ & $570.1 \pm 2.2$\\
Manufacturer & $-896.0 \pm 3.3$ & $-881.2 \pm 4.4$ & $-881.6 \pm 4.7$ & $-880.9 \pm 4.7$ & $\boldsymbol{-880.0 \pm 5.0}$ & $-888.1 \pm 4.4$ & $-863.1 \pm 4.9$ & $-840.7 \pm 5.4$\\
Navigation & $-14.8 \pm 0.4$ & $-13.0 \pm 0.4$ & $-12.9 \pm 0.3$ & $-12.8 \pm 0.3$ & $-12.8 \pm 0.3$ & $-12.8 \pm 0.3$ & $\boldsymbol{-12.5 \pm 0.3}$ & $-12.9 \pm 0.3$\\
Racetrack & $-8.18 \pm 0.02$ & $-8.15 \pm 0.02$ & $-8.16 \pm 0.02$ & $-8.17 \pm 0.02$ & $-8.15 \pm 0.02$ & $-8.07 \pm 0.02$ & $\boldsymbol{-8.07 \pm 0.02}$ & $-8.15 \pm 0.03$\\
Sailing Wind & $-64.6 \pm 1.0$ & $-64.6 \pm 1.4$ & $-64.3 \pm 1.4$ & $-64.6 \pm 1.4$ & $-64.2 \pm 1.4$ & $\boldsymbol{-64.1 \pm 1.5}$ & $-64.5 \pm 1.4$ & $-64.7 \pm 1.3$\\
Skills Teaching & $33.4 \pm 5.6$ & $38.4 \pm 8.0$ & $\boldsymbol{44.9 \pm 7.6}$ & $43.6 \pm 8.0$ & $39.1 \pm 8.4$ & $34.5 \pm 8.7$ & $39.8 \pm 8.3$ & $38.0 \pm 8.4$\\
SysAdmin & $364.2 \pm 1.5$ & $384.8 \pm 2.7$ & $394.8 \pm 2.3$ & $396.1 \pm 2.1$ & $\boldsymbol{396.3 \pm 2.2}$ & $367.1 \pm 2.6$ & $366.8 \pm 2.5$ & $369.9 \pm 2.4$\\
Tamarisk & $-553.3 \pm 8.8$ & $-545.4 \pm 10.0$ & $-548.0 \pm 10.5$ & $-542.6 \pm 10.0$ & $-542.6 \pm 9.8$ & $\boldsymbol{-539.5 \pm 10.3}$ & $-541.3 \pm 9.9$ & $-542.1 \pm 9.8$\\
Traffic & $-10.2 \pm 0.2$ & $-10.0 \pm 0.2$ & $-10.1 \pm 0.3$ & $-10.1 \pm 0.2$ & $-10.3 \pm 0.2$ & $-10.0 \pm 0.3$ & $\boldsymbol{-9.98 \pm 0.24}$ & $-10.2 \pm 0.3$\\
Triangle Tireworld & $68.6 \pm 1.8$ & $74.8 \pm 2.0$ & $74.5 \pm 2.0$ & $74.6 \pm 2.0$ & $\boldsymbol{75.3 \pm 1.9}$ & $69.1 \pm 2.5$ & $72.0 \pm 2.3$ & $74.3 \pm 2.1$\\
\bottomrule
\end{tabular}}

\label{tab:oga_cad_lowvar}
\end{table*}

\noindent \textbf{OGA-CAD drop statistics}: Lastly, we analyze the concrete abstraction drop rate of OGA-CAD for both coarse ($\varepsilon_t,\varepsilon_a$ maximal and partial grouping) and fine abstractions (\mbox{$\varepsilon_t=\varepsilon_a=0$}, no partial grouping) in the above-described low-variance setting. We measure the average ratio of Q nodes of the final search tree that have been dropped and that are part of an abstract node with at least two Q nodes. Tab.\ref{tab:oga_cad_stats} lists these ratios for the first layer, the second layer, and all subsequent layers together.
Clearly OGA-CAD drops coarse abstractions up to $\infty$ times more often than their finer counterparts. The smallest difference by far is in Skills Teaching where OGA-CAD even drops around 5 times more often in the coarse setting. Unsurprisingly, most of the dropping takes place in the first layer as these actions are visited most often.

\begin{table}[]\centering 

\caption{Average ratio of abstraction-dropped Q nodes in OGA-CAD with $p=0.9$ that are part of non-trivial abstractions. The average is denoted for the entire tree, the first, or the second layer only. The columns have the format in which the first entry denotes whether a coarse (C) or fine abstraction has been used (F) and the second entry denotes the layer where T denotes that the entire tree starting at layer 3 has been taken. N/A denotes that no non-trivial abstractions have been constructed in the corresponding layer(s).}

\scalebox{0.75}{\setlength{\tabcolsep}{1mm}\begin{tabular}{c| c c c c c c} \toprule
Domain & C-1 & C-2 & C-T  & F-1 & F-2 & F-T\\ \midrule
Academic Advising & 4.6033\% & 0.6245\%  & 0.3289\% & 0.2778\% & 0.0163\% & 0.0008\% \\ 
Cooperative Recon & 22.278\% & 4.5547\%  & 0.5997\% & 0.3627\% & 0.0995\% & 0.1405\% \\ 
Game of Life & 9.2553\% & 0.1668\%  & 0.0081\% & 0\% & 0\% & N/A \\
Earth of Observation & 47.5502\% & 0.0045\%  & N/A & 0\% & 0\% & N/A \\ 
Manufacturer & 19.6962\% & 0.5647\%  & 0.0984\% & 0\% & 0\% & N/A \\ 
Navigation & 9.8828\% & 7.2123\%  & 3.7305\% & N/A & 0.2389\% & 0.0765\% \\ 
Racetrack & 18.9742\% & 4.6999\%  & 0.0934\% & 0.0704\% & 0.0141\% & 0.0052\% \\ 
Sailing Wind & 10.8386\% & 1.6430\%  & 0.1014\% & 0\% & 0\% & 0.0009\% \\ 
Skills Teaching & 13.24\% & 8.9825\%  & 1.7914\% & 2.7767\% & 2.6417\% & 0.5569\% \\ 
SysAdmin & 5.2672\% & 0.2128\%  & 0.0469\% & 0\% & 0\% & N/A \\ 
Tamarisk & 17.8996\% & 0.1328\%  & 0.0669\% & 0\% & 0.0009\% & 0.0038\% \\ 
Traffic & 12.33\% & 0.2369\%  & 0.0188\% & 0.0162\% & 0\% & N/A \\ 
Triangle Tireworld & 8.0792\% & 8.9461\%  & 3.3704\% & 0.9577\% & 0.2281\% & 0.1403\% \\ \bottomrule
\end{tabular}}   \label{tab:oga_cad_stats} \end{table}

\section{Conclusion and Future Work}
\label{sec:future_work}
Firstly, we introduced OGA-IAAD, an OGA-UCT plugin, to detect whether the abstraction building has no impact on the underlying UCT. OGA-IAAD proved to be a reliable extension that can speedup the OGA-UCT computation whilst not suffering any performance losses. 

Next, we revisited the abstraction-dropping method introduced by Xu et al. \cite{EMCTSXu} and showed that once confounders are eliminated, there are only a handful of instances amongst the considered environments in which this causes a minor performance improvement whilst potentially suffering huge performance losses as OGA-ISD would also drop beneficial abstractions.

As an alternative to OGA-ISD, we proposed OGA-CAD which drops the abstraction on a per-node-basis and only if a certain confidence level is reached that the abstraction is harmful. This confidence-based dropping scheme proved to be very stable, suffering almost no performance losses in all considered settings whilst gaining huge improvements in very coarse (and therefore highly likely erroneous abstractions) abstractions and partly even in very fine abstractions in a low variance Q value setting. We conclude that OGA-CAD when using high confidence values is another beneficial extension to $(\varepsilon_a,\varepsilon_t)$-OGA and also another tool in one's repertoire if one wants to increase the maximal performance in an environment.

As future work, one might investigate how OGA-CAD could be made even more sensitive to coarse abstractions whilst not degrading to a naive dropping scheme to be able to also gain performance boosts in the standard, high-variance setting.

Research into abstraction dropping could also be extended to different abstraction frameworks, such as that of PARSS or even partially-handcrafted frameworks, such as those introduced by Dockhorn et al. \cite{dockAbs}.

Another remaining question is how abstraction dropping can avoid the overestimation problem described in Section \ref{sec:improvements}, as one cannot simply affect the Q value but not the exploration term as that would cause the in-abstraction decision making (i.e. when one state has multiple abstracted actions) to become greedy when all is dropped.

\bibliographystyle{plain}
\bibliography{references}

\newpage 
\onecolumn
\section{Supplementary materials}
\label{sec:appendix}
\subsection{Problem descriptions}
\label{sec:problem_descriptions}
We will provide a brief description of each of the IPPC problems used for the experiments.

\begin{itemize}

    \item \textbf{Sailing Wind}: Originally proposed by Robert Vanderbei \cite{vanderbei1996optimal}, the goal of Sailing Wind is to move a ship that starts at $(1,1)$ on an $n\times n$ grid to $(n,n)$ with minimal cost. There is no consistent use of a transition and reward function throughout the literature. There may just be two available actions (\textit{down}, \textit{right}) \cite{uctJiang} or up to seven (each adjacent cell except the one facing a stochastic wind direction) \cite{AnandGMS15}. The cost of each action is dependent on the current wind direction which stochastically changes its direction at each step independent of the player's actions.

    \item \textbf{Game of Life}: The original game of life by John Conway \cite{gardner1970fantastic} is a cellular automaton and modified into a stochastic MDP as a test problem for the International Probabilistic Planning Competition \cite{sanner2011ippc} by introducing noise to the deterministic state transition, setting the current number of alive cells as the reward, and allowing the agent to choose one cell which will contain a living cell with a high probability. States are elements in $\{0,1\}^{n \times n}$ describing whether there is an alive cell at each cell on a grid. To reduce the action space that scales quadratically which the grid length, we allow only a subset of the original actions, which is to specify one alive cell that is prevented from dying.

    \item \textbf{SysAdmin}: Used as a test problem for the IPPC 2011, a SysAdmin instance is a graph (describing a network topology) with $n \in \mathbb{N}$ vertices. The state space is $\{0,1\}^n$  (describing which machines are currently operating) and the action space is $\{1,\dots,n\}$ (describing with machine to reboot). At each step, the reward is dependent on the machines that are currently working, a reboot causes the rebooted machine to have a high chance of working in the next step. Machines can randomly fail at each step, however this probability is increased when a neighbor fails. 

    \item \textbf{Navigation}: Navigation was a test problem for the International Probabilistic Planning Competition 2011 \cite{sanner2011ippc}. The goal is to move a robot on an $n \times m$ grid from $(n,1)$ to $(n,m)$ in the least number of steps. The robot may move to any of the four adjacent tiles, however, each tile is assigned a unique probability with which the robot is reset back to $(n,1)$. At each step the agent incurs a constant negative reward, making the objective to reach the goal state as quickly as possible.

    \item \textbf{Academic Advising}: The Academic Advising domain was used for the IPPC 2014 \cite{grzes2014ippc}. The agent is a student whose goal is to pass certain academic classes. Formally, the state is an element in $\{H,L,F,NT\}^n$ (representing for each course whether it has not been taken, failed, or passed with a low or high grade) and the agent's action is to choose a course to take. The course outcome depends on the states of the prerequisite courses. The episode ends, when all courses are passed, and while not all mandatory courses, a subset of all courses, are passed, the agent incurs a constant penalty per step.

    \item \textbf{Tamarisk}: Tamarisk is yet another problem from the IPPC 2014 \cite{grzes2014ippc} which models the expansion of an invasive plant in a river system.
    The river system is modelled as a chain of reaches where each reach contains a number of slots that may be unoccupied, occupied by a native plant, or occupied by the invasive Tamarisk plant. Both plant types spread stochastically to neighboring states with a higher probability of spreading downstream. At each time step, the agent chooses an action for one reach, which are doing nothing, eradicating Tamarisk, or restoring a native plant. The action chosen at a reach is applied to all slots in that reach.
    Except for the do-nothing action, all actions can randomly fail. The agent has to balance the action's costs with the penalties incurred for existing Tamarisk plants.

    \item \textbf{Racetrack}: Racetrack was first described by Martin Gardner \cite{gardner1973mathematical} where the goal is to move a car in as few steps as possible to a goal-position on a graph starting from a random position. Each state is a tuple in $\mathbb{Z}^2 \times \mathbb{Z}^2$ that represents the current position and velocity vector. The available actions are adding a vector from $\{-1,0,1\}^2$ to the velocity vector (i.e. accelerating). At each step, with some probability, the chosen acceleration vector is replaced by $(0,0)$. The state transitions by first adding the acceleration vector to the current velocity vector and then adding the velocity vector to the position vector. If a boundary or an obstacle has been hit, the car's position is reset to one of the initial positions with zero velocity. Some implementations however, set the velocity vector to zero and place the car at an empty tile closest to the crash position that is on the straight line connecting the last position to the crash position \cite{OGAUCT}.

    \item \textbf{EarthObservation}: EarthObservation was a test problem for the IPPC 2018 which models a satellite orbiting earth. Formally, each state is a position on a 2-dimensional grid, representing the satellite's longitudinal position and the latitude the camera is aimed at as well as weather levels for some designated cells. At each step, the weather levels stochastically change independent of the agent's actions which are to idle, to take a photo of the current position, or increment/decrement the current cells $y$-position (i.e. shifting the camera focus). A reward is obtained if one of the designated cells is photographed with an amount depending on the cell's current weather condition.

    \item \textbf{Triangle Tireworld:} Tireworld was proposed as a test problem for the IPPC 2004 \cite{YounesLWA05}. In the original goal-based version, the agent is a car that traverses a graph. At each step, the car may move to an adjacent node, change its tire, or load a tire. The goal is to reach a designated goal node. At each step, the car's tire may randomly break. If the car isn't carrying a spare tire, the goal can no longer be reached. Otherwise, if available, a spare tire (at most one can be carried) must be used to replace the current tire. Some nodes contain spare tires, which when the agent visits them, can be picked up.

    \item \textbf{CooperativeRecon}: This domain models a robot having to prove the existence of life on a foreign planet. The robot is modeled as moving on a 2-dimensional grid which contains a number of objects of interest and a base. If the agent is at an object of interest, it can survey the object for the existence of water and life. The probability of a positive result of the latter is dependent on whether water has been detected. If life has been detected, the agent may photograph the object of interest which is the only way to gain a reward. Each detector may break on usage making it either unusable or decreasing its chance of working. The detectors can be repaired at the base.

    \item \textbf{Manufacturer}: In this domain, the agent manages a manufacturing company. The agent's ultimate goal is to sell goods to customers. However, to sell a good, the agent has to first produce the good, which may require building factories and acquiring the necessary goods required for production. Additional difficulty comes from the fact that the goods' price levels vary stochastically.


    \item \textbf{Skills Teaching}: This domain models a student-teacher interaction, where the agent plays the role of the teacher. There is a fixed number of skills that form a directed graph of prerequisites. The student possesses one of three levels of sufficiency at each skill. The agent is rewarded for each skill being at the highest sufficiency and punished for each skill at the lowest sufficiency level. At each, step the agent may choose a skill for which to pose a question to the student or give the student a direct hint. The student can increase their sufficiency at that skill for correctly answering a question and lose sufficiency for answering wrong. The probability of getting a question right is dependent on the sufficiency of the skill's prerequisite. A hint can elevate the student to the medium sufficiency level directly but only if all prerequisites are at the highest sufficiency.

    \item \textbf{Traffic}: This problem models a traffic system in which the agent is tasked with controlling/advancing intersections with the goal of minimizing congestion. The traffic system is modeled as a directed graph and each vertex is either empty or occupied. Occupancy flows along the graph's edges except for some designated intersection edges where the flow is dependent on the intersection's state.
    The only stochasticity of this MDP arises in the form of cars spawning randomly at the designated perimeter vertices. The agent receives a reward equal to the negative number of occupied vertices that have one predecessor vertex that is also occupied.

\end{itemize}

\subsection{$(\varepsilon_a,\varepsilon_t)$-OGA, an extension of OGA to non-exact abstractions}
\label{sec:non_exact_oga}
Preliminary experiments showed that in almost all domains, standard OGA-UCT exclusively detects only action equivalences but abstracts no states. We believe that this is primarily due to two reasons.
\begin{itemize}
    \item 
    Most environments have action spaces featuring at least 10 actions per state. Even when statistically speaking there is a high chance that one action finds a matching action in a potential to-be-grouped state, the probability of finding a match for each action decays exponentially in the number of actions.
    \item Since there is only node at depth zero, state abstractions could only ever be found at depth one of the search tree which themselves have to be bootstrapped of actions abstractions found at depth one. However, due to a high action branching and stochastic branching factor in most environments, the majority of depth one actions have not sampled all possible successors, making finding abstractions difficult.
\end{itemize}
Furthermore, in environments with a high stochastic branching factor, it is practically impossible for two actions to have sampled the exact set of successors, a necessary condition for OGA-UCT. The two above-mentioned problems cause OGA-UCT to miss out on correct abstractions.

Anand et al. \cite{AnandGMS15} partly addressed this problem by introducing pruned OGA-UCT which ignores all successors of state-action pairs whose transition probability $p$ is less than $\alpha \cdot p_{max}$ where $p_{max}$ is the highest transition probability of all sampled successors of the state-action pair in question and $\alpha \in [0,1]$ is some fixed parameter. For this paper however, we investigated the abstraction dropping schemes using $(\varepsilon_a,\varepsilon_t)$-OGA which we will describe now that uses the $(\varepsilon_a,\varepsilon_t)$-ASAP framework. This has three reasons. Firstly, pruned OGA does not allow for errors in the immediate reward. Secondly, pruned OGA is less flexible, as even an alpha value $\alpha=1$ does not guarantee that all states and state-action pairs are grouped. Using the parameters $\varepsilon_a = \infty, \varepsilon_t = 2$, however, guarantees that the coarsest abstraction that simply groups everything, is found. Thirdly, the $\alpha$ value is less interpretable. Even using the value $\alpha=1$ could result in pruned OGA-UCT being equivalent to OGA-UCT in some environments, e.g. those where all transition probabilities are equal. In contrast, the value $\varepsilon_t=0$ means that transition probabilities have to perfectly match and the value $\varepsilon_t=\infty$ implies that the transitions are fully ignored when building the abstraction.

To incorporate the $(\varepsilon_a,\varepsilon_t)$-ASAP framework, we generalized the state-action pair update method as follows as inspired by Jiang et al. \cite{uctJiang} (i.e. the following extension is equivalent to standard OGA-UCT for $\varepsilon_a = \varepsilon_t = 0$). We call the now-to-be-proposed method $(\varepsilon_a,\varepsilon_t)$-OGA. In the following, we say that two state-action pairs are \textbf{similar} if their transition error $F$ is less than or equal to $\varepsilon_t$ and their reward distance is less than or equal to $\varepsilon_a$. We also refer to the maximum of the reward distance and the transition distance as the \textbf{distance} between two state-action pairs. The following modification to the state-action pair update method is a heuristic that balances the stability and size of the abstract nodes

Firstly, each abstract Q node keeps track of its representative which is one of its original Q nodes. Furthermore, it is assigned a unique and constant ID at its creation. At its creation, an abstract node is assigned an ID equal to the total number of abstract nodes that have been created so far.
The update now differentiates between two cases:

\begin{itemize}
    \item 
    \textbf{The state-action pair is its abstract node's representative}: If there is another abstract node to which the current state-action pair is similar and that abstract node is bigger or it is of the same size but has a bigger ID, then we transfer the current action pair to this abstract node. We randomly pick one of the remaining original nodes of the old abstract node as its new representative.

    \item \textbf{State-action pair is not a representative or it's the state-action pair's first update}:
    We check if the state-action pair is no longer similar to its abstract node's representative or if it is the first update call for that node. In that case, we find the abstract node with a similar representative and with minimal distance to the current Q node. If such an abstract node exists, we transfer the current state-action pair to this node. 

\end{itemize}

\subsection{MCTS and OGA-UCT parameters}

\begin{table*}[h]\centering

\caption{A list of the optimal exploration constants for MCTS for both the high variance and low variance setting that we described in Section \ref{sec:dropping_for_improv_exp}. For the high variance setting we tested $\lambda_{highvar} \in \{0.5,1,2,4,8,16,24,32\}$ and for the low variance we tested $\lambda_{lowvar} \in \{0.125,0.25,0.5,1,2,4,8,16,32,64,128,256\}$. By $l_{lowvar}$, we denote the rollout lengths that we used in the low variance setting, and $\varepsilon_a$ lists the immediate reward thresholds that we tested in our experiments. A rollout length of $\infty$ means rollout until the end of the episode. }

\scalebox{1.0}{
\setlength{\tabcolsep}{1mm}\begin{tabular}{c|c c c c}
\toprule
Domain & $\lambda_{highvar}$& $\lambda_{lowvar}$ & $l_{lowvar}$ & $\varepsilon_a$\\
\midrule

SysAdmin & 2 & 4 & 5 & $\{0,1,2\}$\\ 
Game of Life & 1 & 2 & 5 & $\{0,1,2\}$ \\ 
Academic Advising & 4 & 8 & 20 & $\{0\}$ \\ 
Tamarisk & 1 & 1 & 10 & $\{0,0.5,1\}$ \\ 
Cooperative Recon & 2 & 2 & $\infty$ & $\{0,0.5,1\}$ \\ 
Earth Observation & 4 & 64 & 20 & $\{0,1,2\}$ \\ 
Manufacturer & 2 & 2 & 5 & $\{0,10,20\}$ \\ 
Navigation & 8 & 1 & 20 & $\{0\}$ \\ 
Racetrack & 4 & 0.25 & 20 & $\{0\}$ \\ 
Sailing Wind & 2 & 0.5 & 10 & $\{0,1,2\}$ \\ 
Skills Teaching & 4 & 0.5 & 5 & $\{0,2,3\}$ \\ 
Traffic & 1 & 1 & 5 & $\{0,1,2\}$ \\ 
Triangle Tireworld & 2 & 4 & 20 & $\{0\}$ \\ 

\bottomrule
\end{tabular}}

\label{tab:params}
\end{table*}

\end{document}